\documentclass[10pt,twocolumn,letterpaper]{article}

\usepackage{cvpr}
\usepackage{times}
\usepackage{epsfig}
\usepackage{graphicx}
\usepackage{amsmath}
\usepackage{amssymb}
\usepackage{color}
\usepackage{ulem}
\usepackage{booktabs}
\usepackage{multirow, makecell}
\usepackage{nicefrac}

\definecolor{darkgreen}{RGB}{0,150,0}

\definecolor{orange}{RGB}{205,65,0}

\definecolor{Gold}{RGB}{0,204,0}

\def\BG{\mathcal{B}}
\def\CC{\mathcal{C}}
\def\PP{\mathbb{P}}

\usepackage[pagebackref=true,breaklinks=true,letterpaper=true,colorlinks,bookmarks=false]{hyperref}

\cvprfinalcopy 

\def\cvprPaperID{3199} 

\ifcvprfinal\fi
\begin{document}

\title{RepMet: Representative-based metric learning for classification and few-shot object detection}



\newcommand{\printfnsymbol}[1]{$^*$}

\author{Leonid Karlinsky\thanks{The authors have contributed equally to this work}, ~Joseph Shtok\printfnsymbol{1}, Sivan Harary\printfnsymbol{1}, Eli Schwartz\printfnsymbol{1}, Amit Aides, Rogerio Feris\\
IBM Research AI\\
Haifa, Israel \\
\and
Raja Giryes\\
School of Electrical Engineering, Tel-Aviv University\\
Tel-Aviv, Israel\\
\and
Alex M. Bronstein\\
Department of Computer Science, Technion\\
Haifa, Israel\\
}

\maketitle

\begin{abstract}
Distance metric learning (DML) has been successfully applied to object classification, both in the standard regime of rich training data and in the few-shot scenario, where each category is represented by only a few examples. In this work, we propose a new method for DML that simultaneously learns the backbone network parameters, the embedding space, and the multi-modal distribution of each of the training categories in that space, in a single end-to-end training process. Our approach outperforms state-of-the-art methods for DML-based object classification on a variety of standard fine-grained datasets.
Furthermore, we demonstrate the effectiveness of our approach on the problem of few-shot object detection, by incorporating the proposed DML architecture as a classification head into a standard object detection model. We achieve the best results on the ImageNet-LOC dataset compared to strong baselines, when only a few training examples are available. 
We also offer the community a new episodic benchmark based on the ImageNet dataset for the few-shot object detection task. 
Code will be released upon acceptance of the paper.


\end{abstract}

\vspace{-0.2cm}
\section{Introduction}
\label{Introduction}
    
Due to the great success of deep neural networks (DNNs) in the tasks of image classification and detection  \cite{Dai2017b,He2017b,Huang2017, Krizhevsky2012,Simonyan2014,Zoph2017}, they are now widely accepted as the `feature extractors of choice' for almost all computer vision applications, mainly for their ability to learn good features from the data. It is well-known that training a regular DNN model
from scratch requires a significant amount of training data \cite{imagenet}. Yet, in many practical applications, one may be given only a few training samples per class to learn a classifier. This is known as the few-shot learning problem.

\begin{figure}[ht]
\begin{center}
   \includegraphics[width=0.5\textwidth]{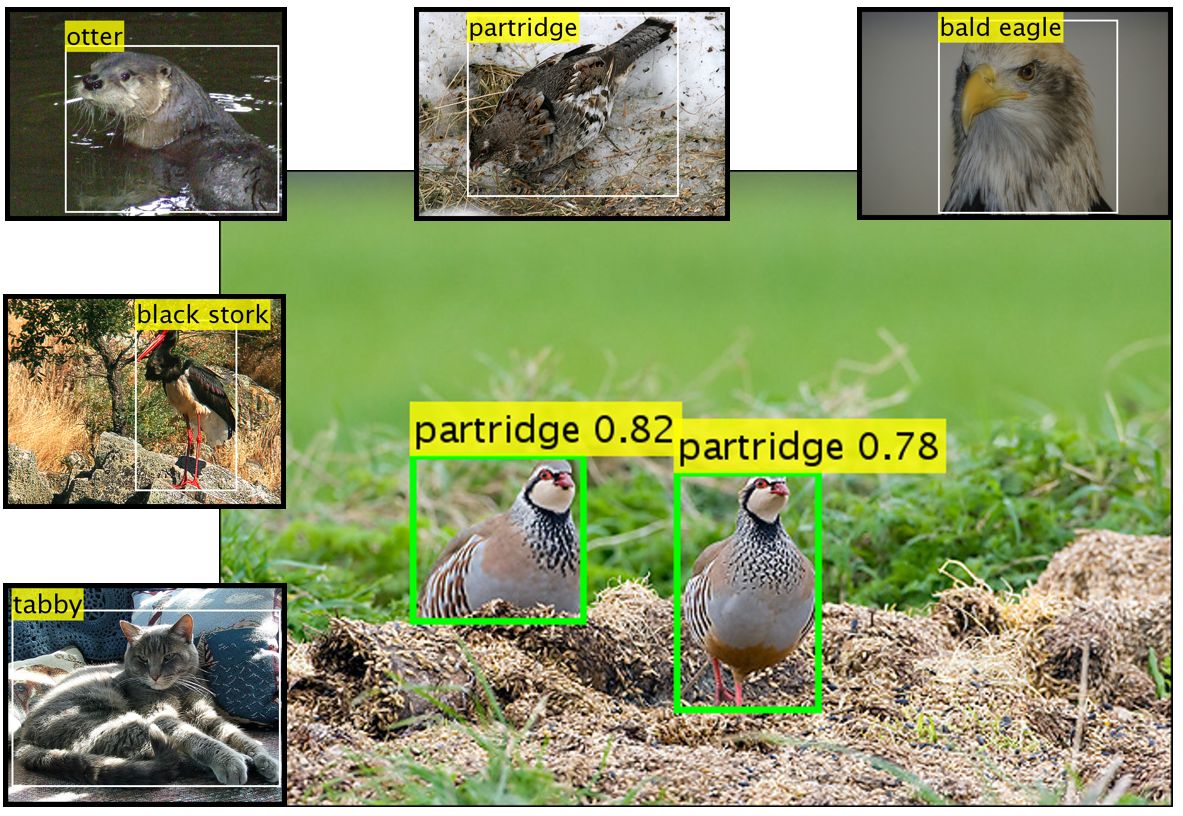}
\end{center}
\vspace{-0.5cm}
   \caption{{\bf One-shot detection example}. Surrounding images: examples of new categories unseen in training. Center image: detection result for the one-shot detector on an image containing instances of {\em partridge}, which is one of the new categories.}
\label{fig:one-shot-test}
\end{figure}

\begin{figure*}[ht]
		\centering
		\includegraphics[width=1.0\textwidth]{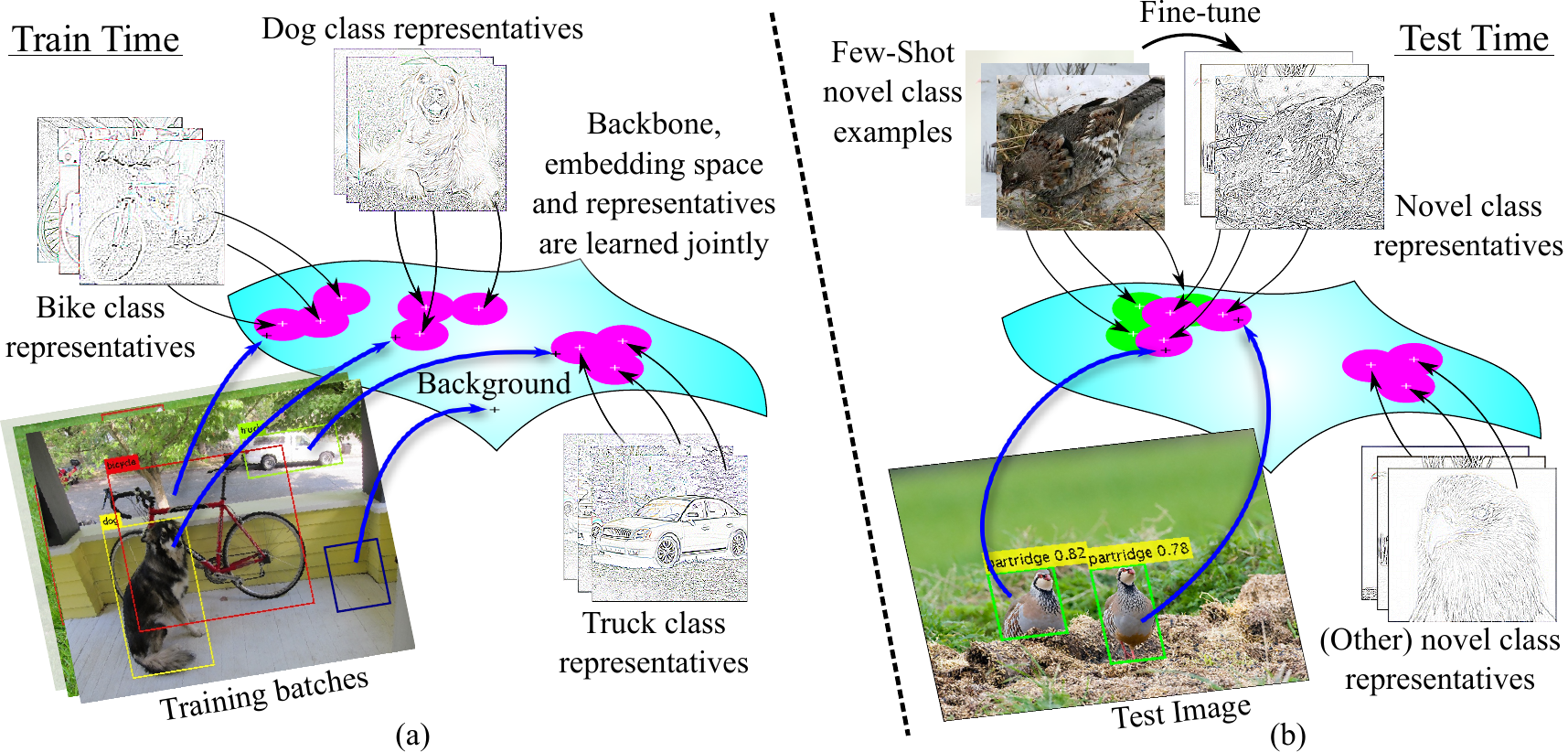}
        \caption{{\bf Overview of our approach.} 
            (a) \textit{Train time:} backbone, embedding space and mixture models for the classes are learned jointly, class representatives are mixture mode centers in the embedding space; (b) \textit{Test time:} new (unseen during training) classes are introduced to the detector in the learned embedding space using just one or a few examples. Fine tuning the representatives and the embedding (on the episode train data) can be used to further improve performance (Section \ref{results}). For brevity, only two novel classes are illustrated in the test. The class posteriors are computed by measuring the distances of the input features to the representatives of each of the classes.
        }\label{fig:one-shot-example}
\end{figure*}

Recent studies have achieved significant advances in using DNNs for few-shot learning. This has been demonstrated for domain-specific tasks, such as face recognition \cite{Schroff2015} and for the classification of general categories
 \cite{Chen2018,Hariharan2017,Snell2017,Vinyals2016,Wang2018,Zhou2018}. However, very few works have investigated the problem of few-shot object \textit{detection}, where the task of recognizing instances of a category, represented by a few examples, is complicated by the presence of the image background and the need to accurately localize the objects. Recently, several interesting papers
 demonstrated preliminary results for the zero-shot object detection case \cite{Bansal2018,Rahman2018} and for the few-shot transfer learning \cite{lstd2018} scenario.
	 
In this work, we propose a novel approach for Distance Metric Learning (DML) and demonstrate its effectiveness on both few-shot object detection and object classification. We represent each class by a mixture model with multiple modes, and consider the centers of these modes as the \textit{representative} vectors for the class. Unlike previous methods, we \textit{simultaneously} learn the embedding space, backbone network parameters, and the representative vectors of the training categories, in a single end-to-end training process.


For few-shot object detection, we build upon modern approaches (e.g., the deformable-FPN variant of the Faster-RCNN \cite{Dai2017b,He2017b}) that rely on a Region Proposal Network (RPN) to generate regions of interest, and a classifier `head' that classifies these ROIs into one of the object categories or a background region.
In order to learn a robust detector with just a few training examples (see Figure \ref{fig:one-shot-test} for a one-shot detection example), we propose to replace the classifier head with a subnet that learns to compute class posteriors for each ROI, using our proposed DML approach.  The input to this subnet are the feature vectors pooled from
the ROIs, and the class posteriors for a given ROI are computed by comparing its embedding vector to the set of representatives for each category. The detection task requires solving `an open set recognition problem', namely to classify ROIs into both the structured foreground categories and the unstructured background category. In this context, the joint end-to-end training is important, since sampling background ROIs for separate training of the DML is very inefficient (Section \ref{results}).

In the few-shot detection experiments, we introduce \textit{new categories} into the detector. This is done by replacing the learned representatives (corresponding to old categories) with embedding vectors computed from the foreground ROIs of the few training examples given for these categories ($k$ examples for $k$-shot detection). We also investigate the effects of fine-tuning our proposed model and the baselines for few-shot learning. Promising results, compared to baselines and the previous work, are reported on the few-shot detection task (Section \ref{results:one-shot-detection}) underlining the effectiveness of jointly optimizing the backbone and the embedding for DML. Figure \ref{fig:one-shot-example} schematically illustrates an overview of our approach to few-shot detection.

%
%
We also demonstrate the use of our approach for general DML-based classification by
comparing to the Magnet Loss \cite{Rippel2015} and other state-of-the-art DML-based approaches \cite{Zhe2018,Qian2015}.
Instead of the alternating training of embedding and clustering used in \cite{Rippel2015}, our proposed approach end-to-end trains a single (monolithic) network architecture capable of learning the DML embedding together with the representatives (modes of the mixture distributions). Effectively, this brings the clustering inside the end-to-end network training. Using this method, we were able to improve upon the state-of-the-art classification results obtained in \cite{Qian2015,Rippel2015,Zhe2018} on a variety of fine-grained classification datasets (Section \ref{results:metric-learning}).

Our contributions are threefold.
%
%
\textbf{First}, we propose a novel sub-net architecture for jointly training an embedding space together with the set of mixture distributions in this space, having one (multi-modal) mixture for each of the categories. 
This architecture is shown to improve the current state of the art for both DML-based object classification and few-shot object detection. 
\textbf{Second}, we propose a method to equip an object detector with a DML classifier head that can admit new categories, and thus transform it into a few-shot detector. To the best of our knowledge, this has not been done before. This is probably due to detector training batches being usually limited to one image per-GPU, not allowing for batch control in terms of category content. This control is needed by any of the current few-shot learners that use episode-based training. This, in turn, makes it challenging to use those approaches within a detector that is being trained end-to-end. In our approach, the set of representatives serves as an `internal memory' to pass information between training batches.
\textbf{Third}, in the few-shot classification literature, it is a common practice to evaluate the approaches by averaging the performance on multiple instances of the few-shot task, called episodes. We offer such an episodic benchmark for the few-shot detection problem, built on a challenging fine-grained few-shot detection task.
    
\section{Related work}
\textbf{Distance Metric Learning.} The use of metric learning for computer vision tasks has a long history (see \cite{Kulis2013} for a survey). 
In a growing body of work, the methods for image classification and retrieval, based on deep DML, have achieved state-of-the-art results on various tasks \cite{Qian2015,Rippel2015,Sohn2016,Zhe2018}. Rippel et al. \cite{Rippel2015} showed that if the embedding and clustering of the category instances are alternated during training, then on a variety of challenging fine-grained datasets \cite{Khosla2011a, Nilsback2008, Parkhi2012, Russakovsky2010} the DML-based classification improves the state-of-the-art even with respect to the non-DML methods.
In DML, the metric being learned is usually implemented as an $L_2$ distance between the samples in an embedding space generated by a neural network.
The basic loss function for training such an embedding is the triplet loss \cite{Weinberger2009}, or one of its recent generalizations \cite{Sohn2016,Song2015a,Wang2017}. These losses are designed to make the embedding space semantically meaningful, such that objects from the same category are close under the $L_2$ distance, and objects from different categories are far apart. This makes DML a natural choice for few-shot visual recognition.
Following the DML, a discriminative class posterior is computed at test time. To that end, a non-parametric approach such as $k$-Nearest-Neighbors ($k$-NN) is commonly used to model the class distributions in the learned embedding space \cite{Vinyals2016,Snell2017,Weinberger2009}, though in some cases parametric models are also used \cite{Chen2012}. In addition, in many approaches such as \cite{Snell2017,Weinberger2009} there is an inherent assumption of the category distributions being uni-modal in the embedding space. Our approach instead learns a multi-modal mixture for each category, while \textit{simultaneously} learning the embedding space in which the distances to these representatives are computed.

\vspace{0.05in}
\textbf{Few-shot Learning.} An important recent work in few-shot classification has introduced Matching Networks \cite{Vinyals2016}, where both train and test data are organized in `episodes'. An $N$-way, $M$-shot episode is an instance of the few-shot task represented by a set of $M$ training examples from each of the $N$ categories, and one query image of an object from one of the categories. The goal is to determine the correct category for the query. In \cite{Vinyals2016}, the algorithm learns to produce a dedicated DML embedding specific to the episode. In \cite{Snell2017}, each class is represented by a Prototype - a centroid of the batch elements corresponding to that category. 
Recently, even more compelling results were obtained on the standard few-shot classification benchmarks using meta-learning methods
\cite{Finn2017,Mishra2017,Ravi2017,Zhou2018} and synthesis methods \cite{Chen2018,Hariharan2017,Schwartz2018,Wang2018,Zhou2018}. 
Although great progress was made towards few-shot classification, it is still difficult to apply these methods to few-shot detection. 
The reason is that a detector training batch typically consists of just one image, with a highly unbalanced foreground to background ROI ratio (somewhat balanced using OHEM \cite{Shrivastava16} and alike). This is problematic for existing few-shot learners, which usually require a balanced set of examples from multiple categories in each batch and commonly have difficulty coping with unstructured noise (background ROIs in our case).

There are only a handful of existing works on few-shot detection. 
An interesting recent work by Chen at al. \cite{lstd2018} proposed using regularized fine-tuning on the few given examples in order to transfer a pre-trained detector to the few-shot task. The authors show that using their proposed regularization, fine-tuning of the standard detectors, such as FRCNN \cite{Ren2015} and SSD \cite{Liu2016b}, can be significantly improved in the few-shot training scenario. 
A different approach by Dong et al. \cite{Dong2017} uses additional unlabeled data in a semi-supervised setting. By using the classical method of enriching the training data with high-confidence sample selection, the method of \cite{Dong2017} produces results comparable to weakly supervised methods with lots of training examples. Unlike previous methods, we propose a DML-based approach for few-shot object detection, which yields superior performance compared to existing techniques.
%
	
\section{RepMet Architecture}\label{section:dml_sub_net}
\begin{figure*}
		\centering
		\includegraphics[width=1.0\textwidth,trim=0cm 4cm 0cm 1.9cm,clip]{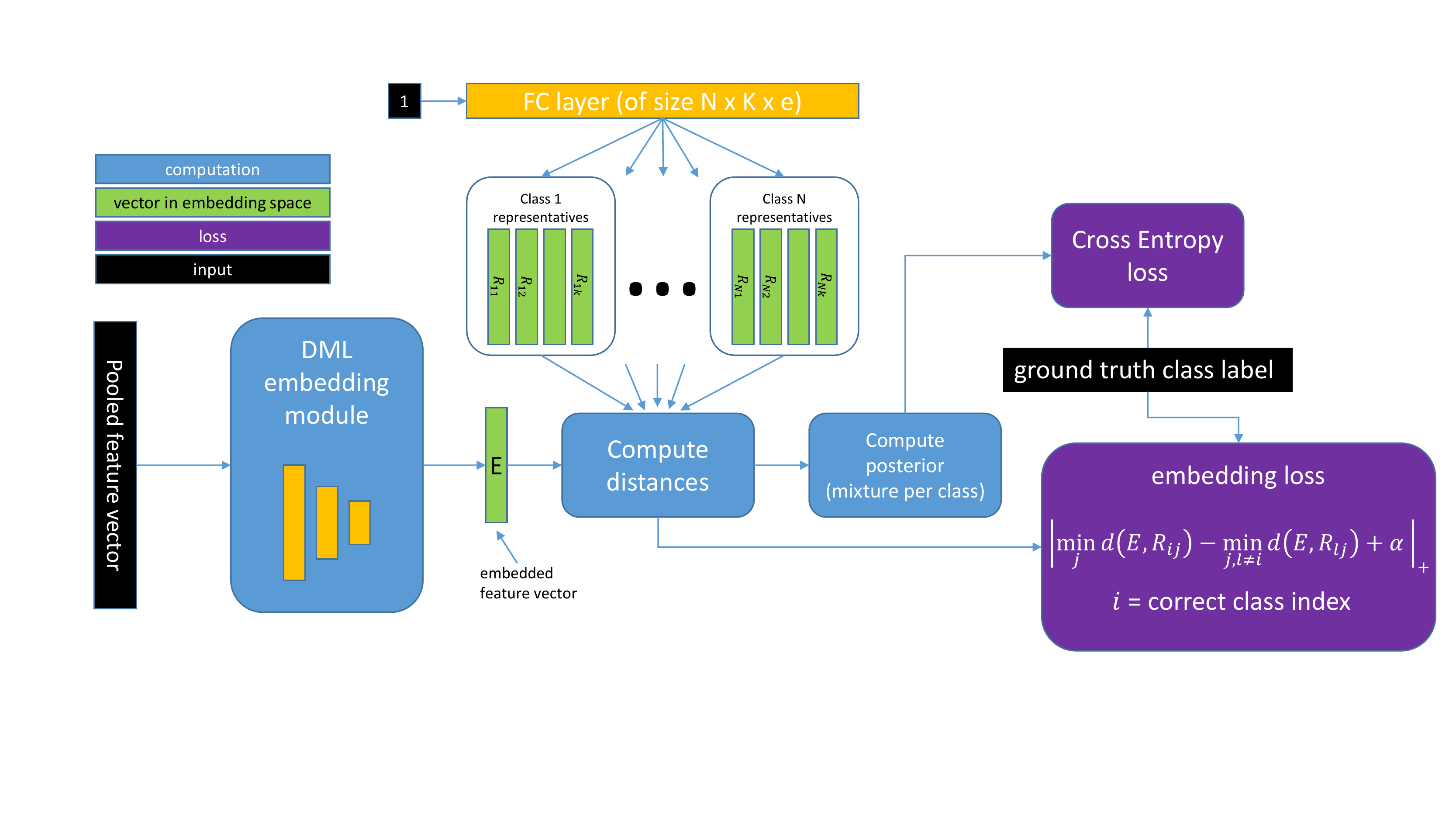}
        \vspace{-0.3cm}
		\caption{{\bf The proposed RepMet DML sub-net architecture} performs joint end-to-end training of the DML embedding together with the modes of the class posterior distribution.}\label{fig:architecture}
\end{figure*}
We propose a subnet architecture and corresponding losses that allow us to train a DML embedding jointly with the multi-modal mixture distribution used for computing the class posterior in the resulting embedding space. This subnet then becomes a DML-based classifier head, which can be attached on top of a classification or a detection backbone. It is important to note that our DML-subnet is trained jointly with the feature producing backbone. The architecture of the proposed subnet is depicted in Figure \ref{fig:architecture}. 

The training is organized in batches, but for simplicity we will refer to the input of the subnet as a single (pooled) feature vector $X \in {\mathbb{R}}^{f}$ computed by the backbone for the given image (or ROI). Examples for a backbone are InceptionV3 \cite{Szegedy2015} or an FPN \cite{Lin2017} (without the RCNN). We first employ a DML embedding module, which consists of a few fully connected (FC) layers with batch normalization (BN) and ReLU non-linearity (we used $2$-$3$ such layers in our experiments). The output of the embedding module is a vector $E=E(X) \in {\mathbb{R}}^{e}$, where commonly $e \ll f$. As an additional set of trained parameters, we hold a set of `representatives' $R_{ij} \in \mathbb{R}^e$. Each vector $R_{ij}$ represents the center of the $j$-th mode of the learned discriminative mixture distribution in the embedding space, for the $i$-th class out of the total of $N$ classes. We assume a fixed
number of $K$ modes (peaks) in the distribution of each class, so $1\leq j\leq K$.
%
%
%

In our implementation, the representatives are realized as weights of an FC layer of size $N \cdot K \cdot e$ receiving a fixed scalar input $1$. The output of this layer is reshaped to an $N \times K \times e$ tensor.
%
%
During training, this simple construction flows the gradients to the weights of the FC layer and learns the representatives.
%
%
For a given image (or an ROI, in the case of the detector) and its corresponding embedding vector $E$, our network computes the $N \times K$ distance matrix whose elements $d_{ij}(E) = d(E,R_{ij})$ are the distances from $E$ to every representative $R_{ij}$. These distances are used to compute the probability of the given image (or ROI) in each mode $j$ of each class $i$:
	\begin{equation}\label{eqn_posterior1}
	{p}_{ij} ( E ) \propto \exp{\left(-\frac{{d^2_{ij}(E)}}{2\sigma^2}\right)}.
	\end{equation}
Here we assume that all the class distributions are mixtures of isotropic multi-variate Gaussians with variance $\sigma^2$. In our current implementation, we do not learn the mixing coefficients and set the discriminative class posterior to be:
	\begin{equation}\label{eq:max_mix}
	\PP(\CC=i|X) = \PP(\CC=i|E) \equiv \max_{j = 1,\dots,K} p_{ij}(E).
	\end{equation}
where $\CC=i$ denotes class $i$ and the maximum is taken over all the modes of its mixture. This conditional probability is an upper bound on the actual class posterior.
The reason for using this approximation is that for one-shot detection, at test time, the representatives are replaced with embedded examples of novel classes, unseen during training (more details are found in Section \ref{results}). Mixture coefficients are associated with specific modes, and since the modes change at test time, learning the mixture coefficients becomes highly non-trivial. Therefore, the use of the upper bound in Eq. \ref{eq:max_mix} eliminates the need to estimate the mixture coefficients. An interesting future extension to our approach would be to predict the mixture coefficients and the covariance of the modes as a function of $E$ or $X$. 

Having computed the class posterior, we also estimate a (discriminative) posterior for the `open' background ($\BG$) class. Following \cite{bendale15}, we do not model the background probability, but instead it is estimated via its lower bound using the foreground (class) probabilities:
	\begin{equation}\label{eqn_backgnd}
	\PP(\BG|X) = \PP(\BG|E) = 1 - \max_{ij} p_{ij}(E).
    \end{equation}\label{eq:bg_prob}
\begin{figure*}
	\centering
	\includegraphics[width=1.0\textwidth,trim=0cm 12.5cm 0cm 1.5cm,clip]{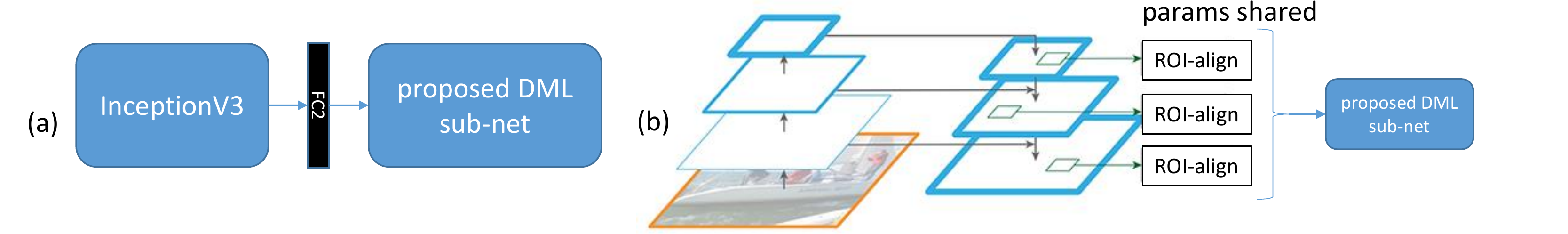}
    \vspace{-0.3cm}
	\caption{{\bf Network architectures used.} (a) Network for DML based classification. (b) Network for few-shot detection; its backbone is FPN+DCN with deformable ROI-align \cite{Dai2017b}.}\label{fig:impl}
\end{figure*}

Having $\PP(\CC=i|X)$ and $\PP(\BG|X)$ computed in the network, we use a sum of two losses to train our model (DML subnet + backbone). The first loss is the regular cross-entropy (CE) with the ground truth labels given for the image (or ROI) corresponding to $X$. The other is intended to ensure there is at least $\alpha$ margin between the distance of $E$ to the closest representative of the correct class, and the distance of $E$ to the closest representative of a wrong class:
\begin{equation}
L(E,R) = \left|\min_j d_{i^\ast j}(E) - \min_{j,i\neq i^\ast} d_{ij}(E) + \alpha\right|_+,
\end{equation}
where $i^\ast$ is the correct class index for the current example and $|\cdot|_+$ is the ReLU function. Figure \ref{fig:impl} illustrates how the proposed DML sub-net is integrated within the full network architectures used for the DML-based classification and the few-shot detection experiments.
	
\section{Implementation details}
In this section we list additional details of our implementation of the proposed approach for the DML-based classification (Section \ref{sec:dml_classification}) and few-shot detection (Section \ref{sec:few_shot_detection}) tasks. Code will be released upon acceptance.

\subsection{DML-based classification}\label{sec:dml_classification}
%
For the DML-based classification experiments, we used the InceptionV3 \cite{Szegedy2015} backbone, attaching the proposed DML subnet to the layer before its last FC layer. The embedding module of the subnet consists of two FC layers of sizes $2048$ and $1024$, the first with BN and ReLU, and the second just with linear activation. 
This is followed by an $L_2$ normalization of the embedding vectors. All layers are initialized randomly. 
%
%
%
In all of our DML-based classification experiments, we set $\sigma = 0.5$ and use $K=3$ representatives per category. Each training batch was constructed by randomly sampling $M=12$ categories and sampling $D=4$ random instances from each of those categories.

In our DML-based classification experiments on standard benchmarks, there is no background category $\BG$, hence we do not need our class mixtures to handle points that are outliers to all of the mixtures. Therefore, we resort to a more classical mixture model variant with equaly weighted modes, replacing the class posterior in Eq. \ref{eq:max_mix} with its softer normalized version, which we have experimentally verified as more beneficial for DML-based classification:
\begin{equation}\label{eq:max_mix_cls}
	\PP(\CC=i|X) = \PP(\CC=i|E) = \dfrac{\sum\limits_{j = 1}^{K} p_{ij}(E) }{     \sum\limits_{i=1}^{N} \sum\limits_{j=1}^{K} p_{ij}(E)
	}
\end{equation}
%
%

\subsection{DML-based few-shot detection}\label{sec:few_shot_detection}
For few-shot detection, we used our DML sub-net instead of the RCNN (the classification `head') on top of the FPN backbone \cite{Lin2017} in its Deformable Convolutions (DCN) variant \cite{Dai2017b}. Our code is based on the original MXNet based implemetation of \cite{Dai2017b}.
The backbone was pre-trained on MS-COCO \cite{Lin2014} from scratch. Our DML subnet, including the representatives, was initialized randomly. The entire network was trained in an end-to-end fashion using OHEM \cite{Shrivastava16} and SoftNMS \cite{Bodla2017}.
The embedding module in the DML subnet for one-shot detection consisted of two FC layers of width $1024$ with BN and ReLU, and a final FC layer of width $256$ with linear activation, followed by $L_2$ normalization. We used $K=5$ representatives per class during training, and set $\sigma = 0.5$. As in \cite{Dai2017b}, each training batch consisted of one random training image.
	
\section{Results}
\label{results}
We have evaluated the utility of our proposed DML subnet on a series of classification and few-shot detection tasks.
\subsection{DML-based classification}\label{results:metric-learning}
\begin{figure*}[h]
	\centering
	\includegraphics[width=0.75\textwidth,trim=0cm 10.5cm 4cm 0cm,clip]{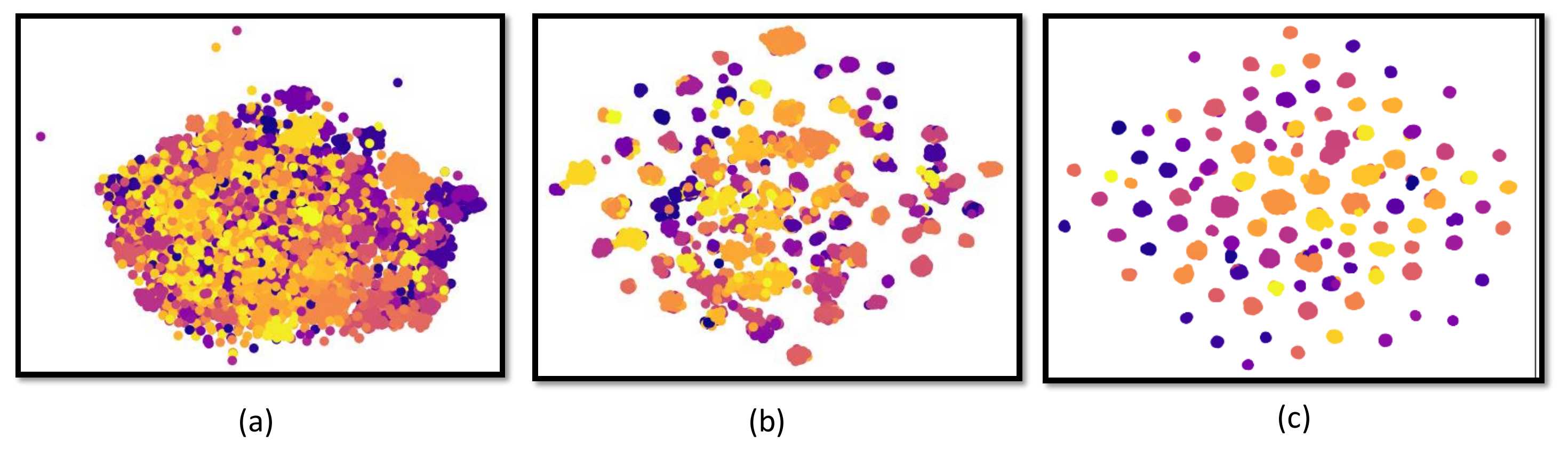}
	\caption{Evolution of the t-SNE visualization of the  embedding space while training on the Oxford Flowers. Different colors correspond to different mixture modes. (a) initial; (b) 1200 iterations; (c) 4200 iterations. Note the separation into local clusters created by our DML.}
	\label{fig:results_dml_tsne}
\end{figure*}

%
\textbf{Fine-grained classification.} We tested our approach on a set of fine-grained classification datasets, widely adopted in the state-of-the-art DML classification works \cite{Qian2015,Rippel2015,Zhe2018}: Stanford Dogs \cite{Khosla2011a}, Oxford-IIIT Pet \cite{Parkhi2012}, Oxford 102 Flowers \cite{Nilsback2008}, and ImageNet Attributes \cite{Rippel2015}. The results reported in Table \ref{table:Magnet_perf} show that our approach outperforms the state-of-the-art DML classification methods \cite{Qian2015,Rippel2015,Zhe2018} on all datasets except Oxford Flowers. Figure \ref{fig:results_dml_tsne} shows the evolution of the t-SNE \cite{VanDerMaaten2008} plot of 
the training instances in the embedding space over the training iterations.

\begin{table*}[h]
	\centering
	\begin{tabular}{lcccc}
		\toprule
    & \multicolumn{4}{c}{method} \\
    \cmidrule(lr){2-5}
    dataset & MsML \cite{Qian2015}  & Magnet \cite{Rippel2015}  & VMF \cite{Zhe2018} & Ours \\
		\midrule
		Stanford Dogs       &  29.7  & 24.9 & 24.0          &  $\textbf{13.7}$    \\
		Oxford Flowers      & 10.5   & 8.6  & $\textbf{4.4}$  & 11  \\
		Oxford Pet          & 18.8   & 10.6 & 9.9           &$\textbf{ 6.9}$ \\
		ImageNet Attributes	& --     & 15.9	& ---           & $\textbf{13.2}$  \\		
		\bottomrule
	\end{tabular}
    \caption{Comparison of \textbf{test error (in \%)} with the state-of-the-art DML classifier approaches on different fine-grained classification datasets (lower is better).}
    \label{table:Magnet_perf}
\end{table*}

\begin{figure*}[ht]
	\centering
	\includegraphics[width=0.75\textwidth,trim=2cm 9cm 3cm 0cm,clip]{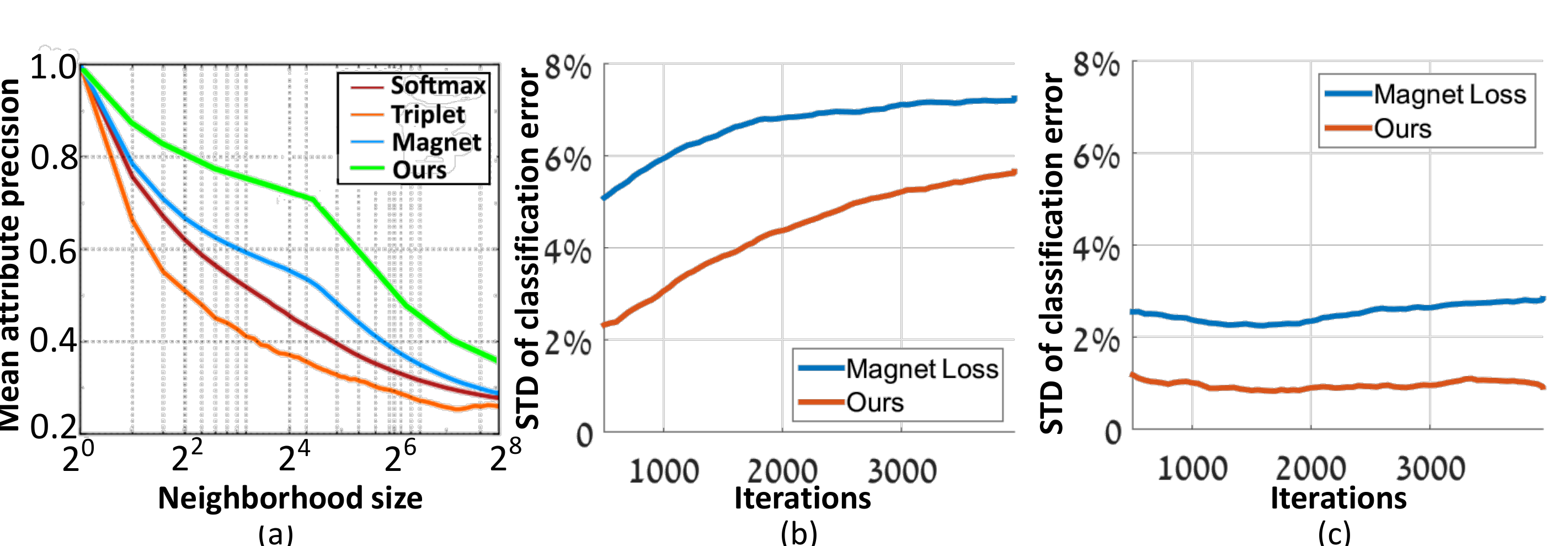}
	\caption{(a) Mean attribute precision as a function of neighborhood size on the ImageNet Attributes dataset. The `Softmax', `Triplet' and `Magnet' graphs are borrowed from \cite{Rippel2015}. (b) Testing performance stability to hyperparameter change of our method and the Magnet loss \cite{Rippel2015}. We plot the STD of the classification error, measured accross various depth and width sizes of the embedding model, as a function of the iteration number. Lower is better. 
	(c) same as (b) for various number of modes in the learned mixture.}
	\label{fig:results_graph1}
\end{figure*}

\textbf{Attribute distribution.} We verified that following DML training for classification, images with similar attributes are closer to each other in the embedding space (even though attribute annotations were not used during training). We used the same experimental protocol as \cite{Rippel2015}. Specifically, we trained our DML classifier on the ImageNet Attributes dataset defined in \cite{Rippel2015}, which contains $116236$ images from $90$ classes. Next, we measured the attribute distribution on the Object Attributes dataset \cite{Russakovsky2010}, which provides $25$ attributes annotations for about $25$ images per class for these $90$ classes.
For each image in this dataset, and for each attribute, we compute the fraction of neighbors also featuring this attribute, over different neighborhood cardinalities. Figure \ref{fig:results_graph1}(a) shows improved results obtained by our approach as compared to \cite{Rippel2015} and to other methods.

\textbf{Hyperparameter robustness -- ablation study.} We evaluated different values of representatives per class ($1 \le K \le 8$), and $9$ different architectures of the embedding network (varying the number of dense layers between $1$ and $3$ and using three different widths for each). 
Same robustness tests were also repeated for our implementation of \cite{Rippel2015}. We have verified that our implementation reproduced the results reported in \cite{Rippel2015}, whose code is not available.

Figures \ref{fig:results_graph1}(b) and \ref{fig:results_graph1}(c) show that our method is more robust to changes in the hyperparameters compared to \cite{Rippel2015}. These figures depict, for each method and at each training iteration, the standard deviation of the classification error obtained by varying the embedding network architecture and the number of representatives per class, respectively.

\subsection{Few-shot object detection}
\label{results:one-shot-detection}


To the best of our knowledge, the only few-shot detection benchmark available to-date is reported in the LSTD work \cite{lstd2018} by Chen et al., who proposed to approach few-shot detection by a regularized fine-tuning. In Table \ref{table:comp2lstd2018}, we compare our approach to the results of LSTD \cite{lstd2018} on 'Task 1', which is their most challenging ImageNet based $50$-way few-shot detection scenario. 
\begin{table}[h]
	\small
	\centering
	\begin{tabular}{lccc}
		\toprule
    & 1-shot & 5-shot & 10-shot\\
		\midrule
		LSTD \cite{lstd2018}  & 19.2 & 37.4 & 44.3 \\
		ours & \textbf{24.1} & \textbf{39.6} & \textbf{49.2} \\
		\bottomrule		
	\end{tabular}		
	\caption{ Comparison to LSTD \cite{lstd2018} on their Task 1 experiment: $50$-way detection on 50 ImageNet categories (as mAP $\%$). }
    \label{table:comp2lstd2018}
\end{table}

%

\begin{table*}
  \small \centering
  \begin{tabular}{lccccccc}
    \toprule
    & & \multicolumn{3}{c}{no episode fine-tuning} & \multicolumn{3}{c}{with episode fine-tuning} \\
    \cmidrule(lr){3-5}\cmidrule(lr){6-8}
     dataset  & method & 1-shot & 5-shot & 10-shot & 1-shot & 5-shot & 10-shot \\
    \midrule
    \multirowcell{3}[0pt][l]{\textbf{ImageNet-LOC} \\ (214 unseen animal classes)} & baseline-FT (FPN-DCN \cite{Dai2017b})       & --- & --- & --- & 35.0 & 51.0 & 59.7 \\
     & baseline-DML       & 41.3 & 58.2 & 61.6 & 41.3 & 59.7 & 66.5 \\        
     & baseline-DML-external       & 19.0 & 30.2 & 30.4 & 32.1 & 37.2 & 38.1 \\        
     & Ours    		 & \textbf{56.9} & \textbf{68.8} & \textbf{71.5} & \textbf{59.2} & \textbf{73.9} & \textbf{79.2} \\
    \midrule
    \multirowcell{2}[0pt][l]{\textbf{ImageNet-LOC} \\ (100 seen animal classes)} & Ours - trained representatives   & --- & 86.3 & --- & --- & --- & --- \\
     & Ours - episode representatives   & 64.5 & 79.4 & 82.6 & --- & --- & --- \\
    \bottomrule
  \end{tabular}
  \caption{Few-shot $5$-way detection test performance on ImageNet-LOC. Reported as mAP in \%.}
  \label{table:results_fsd_1}
\end{table*}

%
%

Since for all of their proposed tasks, the benchmarks of \cite{lstd2018} consist of just one episode (train/test images selection) per task, we created an additional benchmark for few-shot detection. Our proposed benchmark is based on ImageNet-LOC data. The benchmark contains multiple random episodes (instances of the few-shot detection tasks); we used $500$ random episodes in our benchmark. This format is borrowed from the few-shot classification literature.
Each episode, for the case of the $n$-shot, $m$-way few-shot detection task, contains random $n$ training examples for each of the $m$ randomly chosen classes, and $10 \cdot m$ random query images containing one or more instances belonging to these classes (thus at least $10$ instances per class). The goal is to detect and correctly classify these instances.
For consistency, for each $n\in \{1,5,10\}$ the same $500$ random episodes are used in all of the $n$-shot experiments.
Please see Figure \ref{fig:one-shot-test} for an illustration of a $1$-shot, $5$-way episode. 

On the proposed few-shot detection benchmark, we have compared our approach to three baselines. For the first, denoted as \textbf{`baseline-FT'}, we fine-tune a standard detector network on just the few ($n \cdot m$) available samples of the ($m$) novel categories in each ($n$-shot, $m$-way) test episode. Specifically, we fine-tuned the linear decision layer of the classifier head of the FPN-DCN detector \cite{Dai2017b}, the same detector we use as a backbone for our approach.
%
%
For the second baseline, denoted as \textbf{'baseline-DML'}, we
attach our DML sub-net without the embedding module to the regular (pre-trained) FPN-DCN detector, effectively using the FPN-DCN two last FC layers as the embedding module. The FPN-DCN detector used for this baseline is pre-trained as a regular FPN-DCN on the same data as our approach, hence without being optimized for DML based classification as opposed to our full method. 
%
%
%
For the third baseline, denoted as \textbf{`baseline-DML-external'}, we trained the DML sub-net embedding module separately from the detector, in an offline training process. The embedding was trained on sampled foreground and background ROIs using the triplet loss \cite{Weinberger2009}. We also obtained a similar performance for this baseline when training the embedding module with the Prototypical Networks \cite{Snell2017}.
%

All the baselines were pre-trained on the same training set as our model and tested on the same collections or random episodes. To train the models we used 
the $100$ first categories from ImageNet-LOC (mostly animals and birds species).
For testing, we used all the remaining $214$ ImageNet-LOC animal and bird species categories (unseen at training) to ensure that the train and the test categories belonged to the same concept domain. For our model and all the DML-baselines, in each episode, the set of categories being detected was reset to the $m$ new ones by replacing the set of representatives $R$ in the DML subnet with the embedding vectors computed from the ROIs corresponding to the training objects of the episode. These ROIs were selected among the $2K$ ROIs per image returned by RPN by checking which ROIs passed the IoU$\ge 0.7$ requirement with the training objects bounding boxes. In our approach, the embedding and the backbone are jointly optimized to be used with the representatives-based class posterior. This offers an advantage compared to the baselines, as suggested by the performance comparison reported in Table \ref{table:results_fsd_1}.

    \begin{figure*}[h]
	\centering
	\includegraphics[width=1.0\textwidth,trim=0cm 0cm 0cm 0cm,clip]{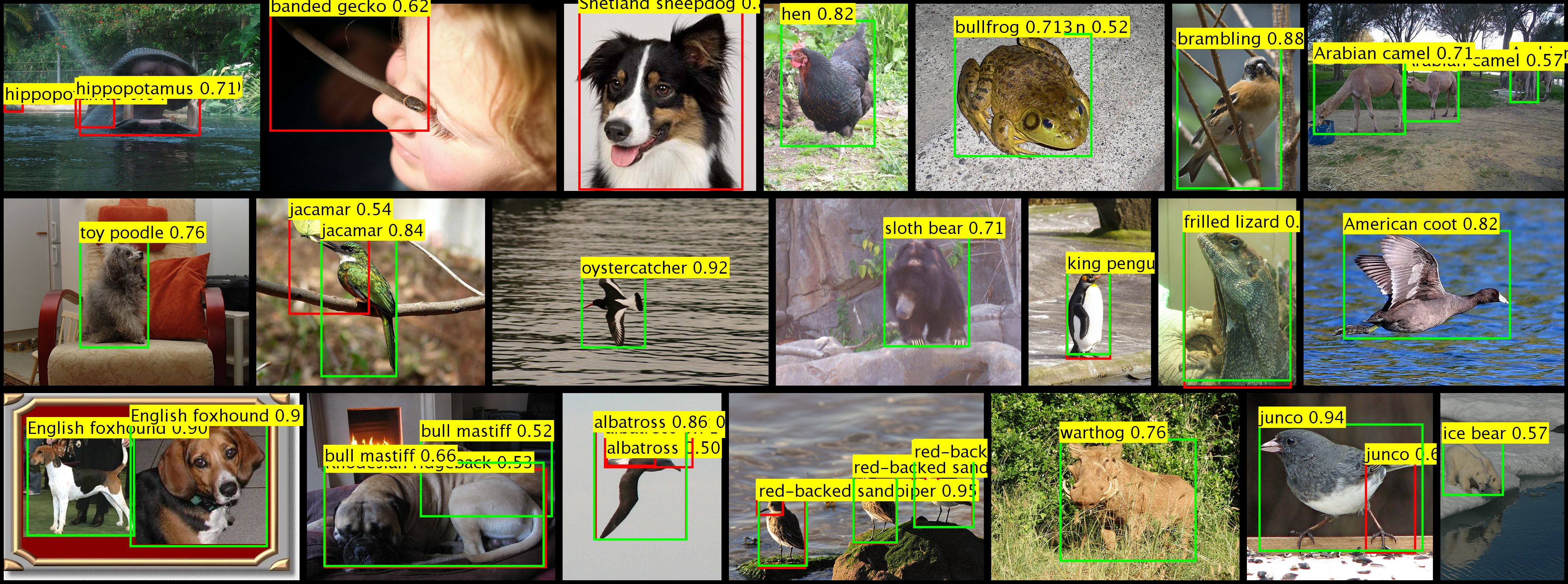}
	\caption{{\bf Example one-shot detection results.} Green frames indicate correctly detected objects and red frames indicate wrong detections. A threshold of $0.5$ on the detection score is used throughout. Detections with higher scores are drawn on top of those with lower scores.}
	\label{fig:five-shot-visual}
\end{figure*}

The evaluation of our approach and the baselines on the set of unseen classes is reported in Table \ref{table:results_fsd_1} (in its \emph{unseen classes} section). The mean average precision (mAP) in \% is calculated on $5$-way detection tasks ($500$ such tasks). The mAP is computed by collecting and evaluating jointly (in terms of score threshold for computing precision and recall) the entire set of bounding boxes detected in all the $500$ test episodes with $50$ query images each. 

In addition, for each of the tested methods (ours and the baselines), we repeated the experiments while fine-tuning the last layer of the network just on the episode training images (for our model and the baselines using DML, the last embedding layer and the representatives were fine-tuned). The results with fine-tuning are also reported in Table \ref{table:results_fsd_1}.
Figure \ref{fig:five-shot-visual} shows examples of $1$-shot detection test results.

%
%
%

From the relatively low performance of 'baseline-DML-external', we can conclude that, as stated in the introduction, joint training of the embedding space with the DML classifier is crucial for the performance. From our close examination, the reduction in mAP of 'baseline-DML-external' is mostly attributed to significantly higher False Positives rates than in the other methods. Although the external embedding was trained on the same training images as our method and the other baselines, it was infeasible to sample the entire collection of possible background ROIs that are being processed by our method when training as a detector end-to-end. Therefore, we had to resort to sampling 200 ROIs per image, which reduced the baseline's ability to reject the background.

To test the inter-dependence of the learned embedding on the specific representatives learned jointly with it during training, we repeated the episode-based testing on the set of classes seen during training (using only validation images not used for training). The results of this evaluation are also reported in Table \ref{table:results_fsd_1} in the \emph{seen classes} section. We repeated the seen classes testing twice: once using the representatives taken from the training objects of each episode (same as for unseen classes) and once using the originally trained representatives (as they correspond to the set of seen classes). Since during training, we learn $K=5$ representatives per class, we report the result of the second test in the $5$-shot column. We can see that (i) the trained representatives perform better than embedding of random class examples, underlining again the benefits of joint training; (ii) the performance drop from trained representatives to random class members is not that big ($\sim7$ points), hinting that the learned embedding is robust to change of representatives and is likely to perform well on the new unseen categories (as was verified above in our few-shot experiments). 

In \cite{Bansal2018} Recall@100 was used as their performance measure (Recall \% taking $100$ top detections in each test image). We also implemented this measure in our $1$-shot test, achieving 88.2\% Recall@100  and 65.9\% Recall@10 calculated over our entire set of $500$ test episodes. This demonstrates that our approach works well on an individual image basis, and illustrates the importance of considering all the boxes from all the test images simultaneously when computing the AP, as we did in our benchmark.

In order to check if the modification introduced by replacing the RCNN classifier with our DML sub-net hinders the detection performance on the seen classes, we tested the detection performance of our model and the vanilla FPN-DCN model (using their original code) on the validation sets of the 100 first Imagenet-LOC training categories and of PASCAL VOC. As shown in Table \ref{table:results_fsd_2}, our detector is slightly inferior to the original FPN-DCN model on the Pascal VOC, but compares favorably on the 100 first Imagenet-LOC (more fine-grained) categories.
\begin{table}[ht]
	\small
	\centering
	\begin{tabular}{lcccccc}
		\toprule
    & \multicolumn{3}{c}{PASCAL VOC}  & \multicolumn{3}{c}{ImageNet (LOC)}  \\
    \cmidrule(lr){2-4}\cmidrule(lr){5-7}
    method / IoU & 0.7 & 0.5 & 0.3 & 0.7 & 0.5 & 0.3 \\
		\midrule
		FPN-DCN \cite{Dai2017b}  & \textbf{74.6} & \textbf{83.5} & \textbf{85.3} & 46.9 & 55.2 & 60.2      \\
		Ours    & 73.7 & 82.9 & 84.9 & \textbf{60.7} & \textbf{61.7} & \textbf{70.7}  \\
		\bottomrule
	\end{tabular}
	\caption{Regular detection performance (in mAP [\%]) per different acceptance
    IoU. FPN-DCN evaluated using their original code.}
    \label{table:results_fsd_2}
\end{table}


\section{Summary \& Conclusions}\label{summary}

In this work, we proposed a new method for DML, achieving state-of-the-art performance for object classification compared to other DML-based approaches. Using this method, we designed one of the first few-shot detection approaches, which compares favorably to the current few-shot detection state-of-the-art. We also proposed a benchmark for the few-shot object detection, based on the Imagenet-LOC dataset, in the hopes that it will encourage researchers to further investigate into this problem, which has so far been almost untouched.
Future work directions include predicting the mixing coefficients and covariances for the class mixtures learned within our DML sub-net as a function of the input. We also intend to use example synthesis methods to generate new samples on-the-fly automatically, thereby enriching the set of estimated representatives.

{\small
\bibliographystyle{ieee}
\bibliography{Mendeley_OR/DL_team_reading_group}
}

\end{document}